\documentclass[runningheads]{llncs}

\usepackage[T1]{fontenc}

\usepackage{graphicx}
\usepackage{booktabs}

\usepackage{hyperref}
\usepackage{subcaption}

\usepackage{orcidlink}

\usepackage{amsmath,amssymb}
\usepackage{algorithm}
\usepackage{algpseudocode}
\usepackage{wrapfig}

\newcommand{\Real}{{\mathbb{R}}}

\begin{document}

\title{Box2Flow: Instance-based Action Flow Graphs from Videos} 


\author{Jiatong Li\inst{1} \and
Kalliopi Basioti\inst{1} \and
Vladimir Pavlovic\inst{1}}

\authorrunning{J.~Li et al.}

\institute{Rutgers University, Piscataway NJ 08854, USA}

\maketitle

\begin{abstract}
  A large amount of procedural videos on the web show how to complete various tasks. These tasks can often be accomplished in different ways and step orderings, with some steps able to be performed simultaneously, while others are constrained to be completed in a specific order. 
  Flow graphs can be used to illustrate the step relationships of a task. 
  Current task-based methods try to learn a single flow graph for all available videos of a specific task. 
  The extracted flow graphs tend to be too abstract, failing to capture detailed step descriptions. 
  In this work, our aim is to learn accurate and rich flow graphs by extracting them from a single video. We propose \texttt{Box2Flow}, an instance-based method to predict a step flow graph from a given procedural video. In detail, we extract bounding boxes from videos, predict pairwise edge probabilities between step pairs, and build the flow graph with a spanning tree algorithm. Experiments on MM-ReS and YouCookII show our method can extract flow graphs effectively.

  
\end{abstract}

\section{Introduction}
\vspace{-.5\baselineskip}
\label{sec:intro}
Procedural videos showing how to perform various tasks can be found on video sharing platforms, ranging from adding oil to cars to making cakes. This wealth of data creates an opportunity for computer vision systems \cite{ZhXuCoAAAI18,zhukov2019cross,zhou2023procedure} to learn a computational representation of those multi-step procedures, which can then be used in various downstream applications ranging from video activity segmentation to general procedure analytics. 

\begin{figure}[t]
    \centering
    \begin{subfigure}[t]{0.65\textwidth}
        \centering
        \includegraphics[width=1.0\textwidth]{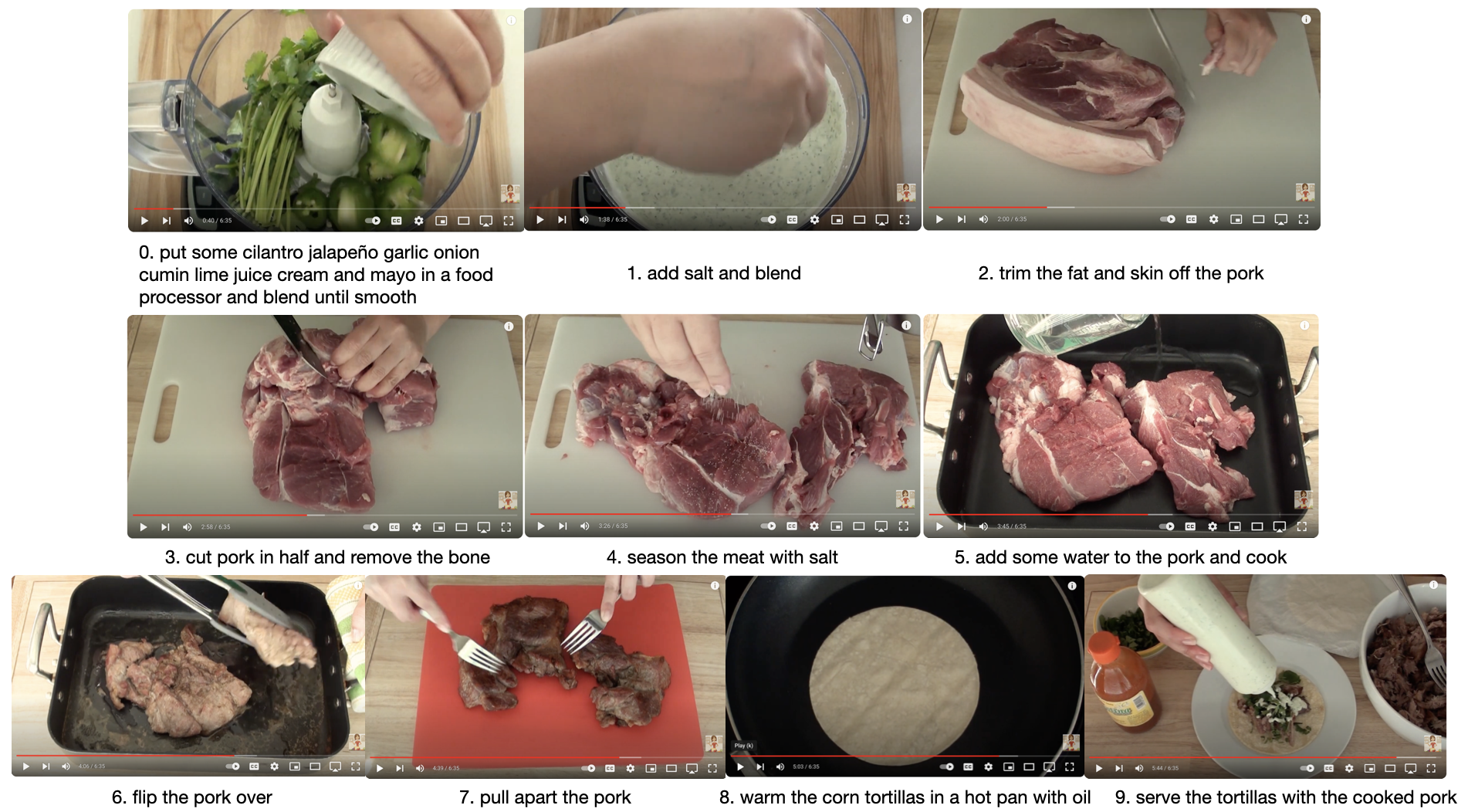}
        \vspace{-.5\baselineskip}
        \caption{\centering \small Carnitas Tacos With Cilantro Lime Sauce.}
        \label{fig:recipe_intro1}
    \end{subfigure}
    \begin{subfigure}[t]{0.3\linewidth}
        \centering
        \includegraphics[width=0.9\textwidth]{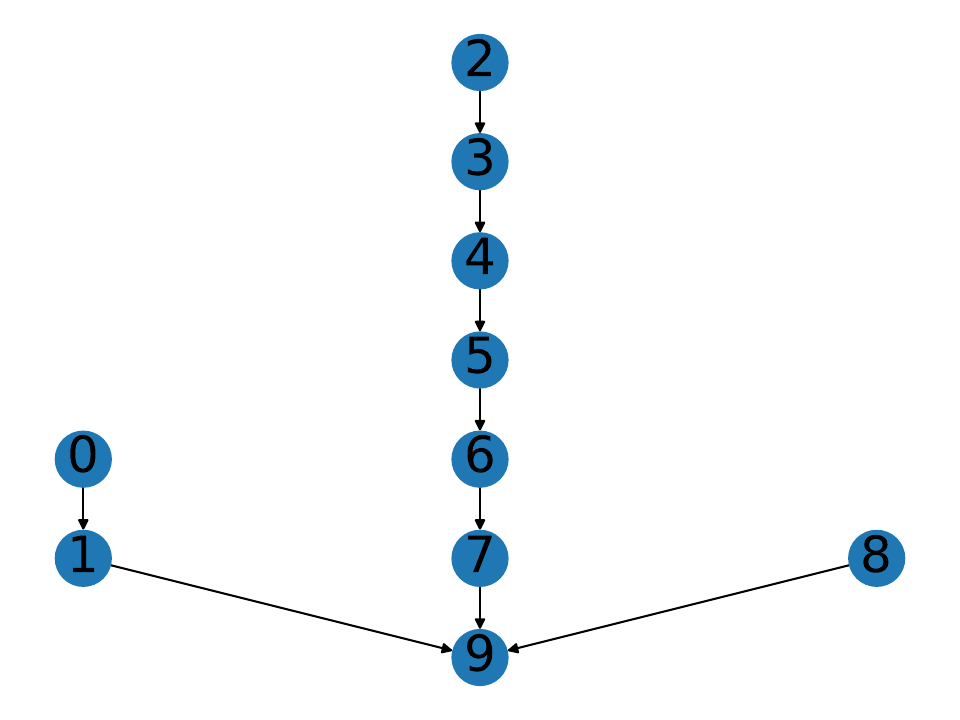}
        \vspace{-.5\baselineskip}
        \caption{\centering \small Flow Graph of Carnitas Tacos With Cilantro Lime Sauce.}
        \label{fig:example_flow}
    \end{subfigure}
    \begin{subfigure}[t]{0.65\linewidth}
        \centering
        \includegraphics[width=1.0\textwidth]{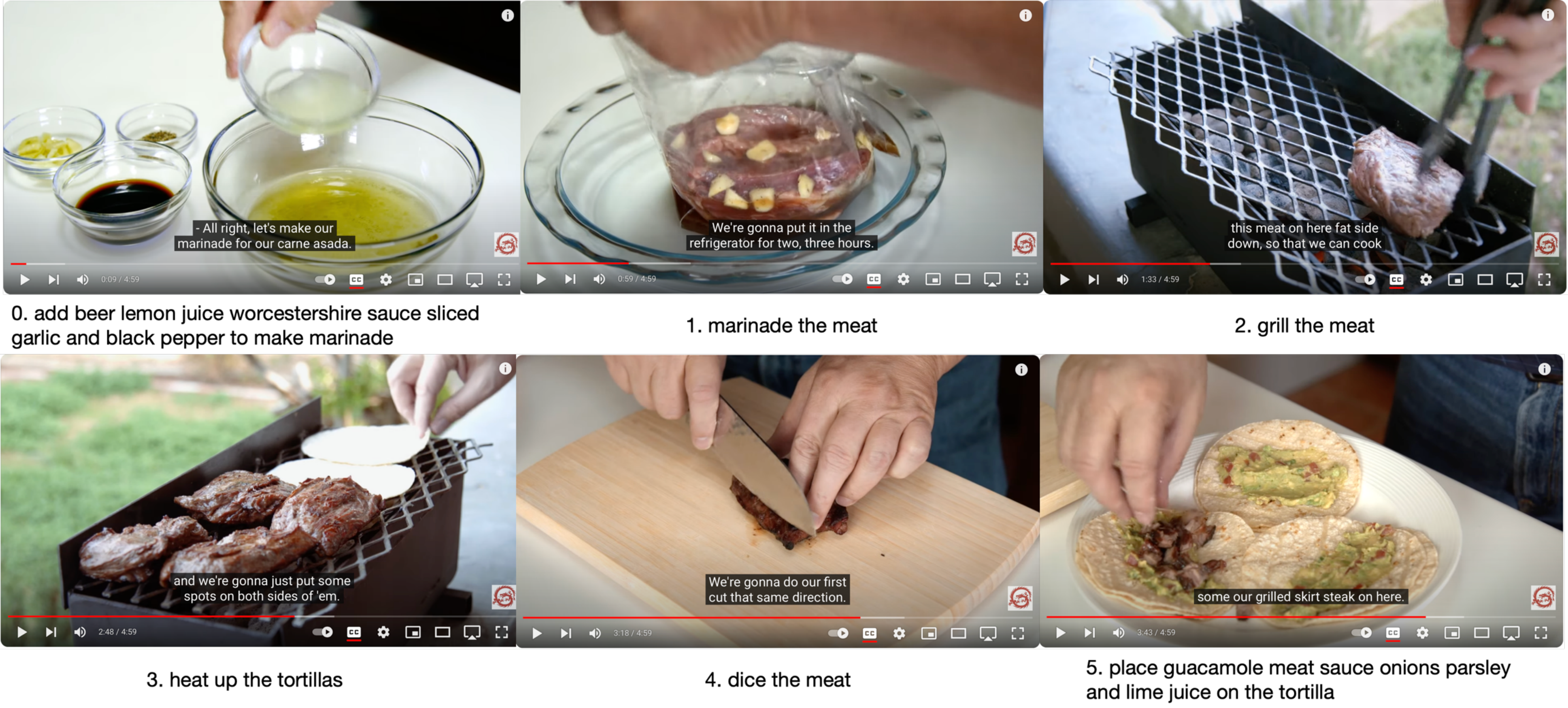}
        \vspace{-.5\baselineskip}
        \caption{\centering \small Carne Asada Tacos.}
        \label{fig:recipe_intro2}
    \end{subfigure}
    \begin{subfigure}[t]{0.3\linewidth}
        \centering
        \includegraphics[width=0.9\textwidth]{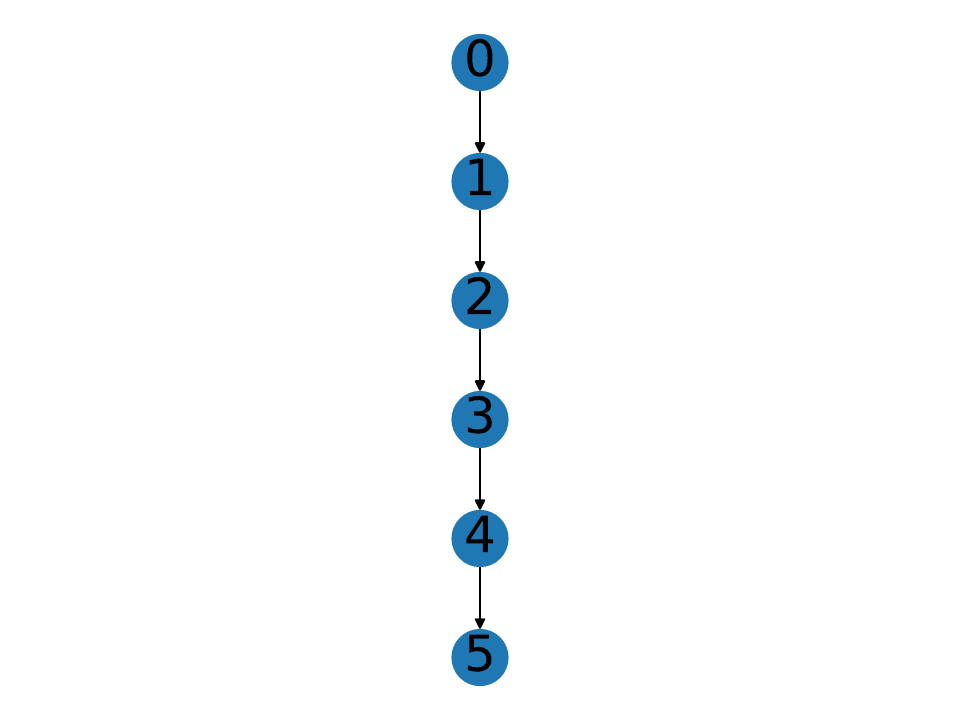}
        \vspace{-.5\baselineskip}
        \caption{\centering \small Flow Graph of Carne Asada Tacos.}
        \label{fig:example_flow2}
    \end{subfigure}
    \vspace{-1\baselineskip}
    \caption{ (a), (c): Two recipes for making tacos that differ in ingredients, actions, and number of steps. (b), (d): The corresponding flow graphs of the two recipes.}
    \vspace{-1\baselineskip}
\end{figure}

However, in real-world procedures seemingly identical tasks are often performed differently by individual users, including, e.g., using different materials or cooking ingredients, different actions, different step orderings, and different number of steps, while also sharing some common procedural elements. This will lead to different procedure workflows depending on each instance of the task as recorded in a video. As shown in Figure \ref{fig:recipe_intro1}, \ref{fig:recipe_intro2}, two recipes, \texttt{Carnitas Tacos With Cilantro Lime Sauce} and \texttt{Carne Asada Tacos}, both belong to the same task category, making tacos, but they are very different. The two recipes use different ingredients for flavoring. \texttt{Carnitas Tacos With Cilantro Lime Sauce} added the sauce in the last step while \texttt{Carne Asada Tacos} marinated the meat. In terms of actions, \texttt{Carnitas Tacos With Cilantro Lime Sauce} cooked the pork with water and pulled apart the pork while \texttt{Carne Asada Tacos} grilled and diced the meat. \texttt{Carnitas Tacos With Cilantro Lime Sauce} heated the tortillas after pulling apart the pork while \texttt{Carne Asada Tacos} heated the tortillas before dicing the meat. Finally, \texttt{Carnitas Tacos With Cilantro Lime Sauce} is annotated with 10 steps while \texttt{Carne Asada Tacos} is annotated with 6. As a result, their workflows are also different, as in Figure \ref{fig:example_flow}, \ref{fig:example_flow2}. 
Therefore, in order to comprehend procedural videos, a computer vision system must be able to recognize the various types of steps and their possible sequences. We propose that disassembling each video instance separately into individual steps can lead to a better understanding of the overall task than task-based methods \cite{zhou2023procedure,jang2023multimodal,mao2023action} that try to learn task steps simultaneously by processing all available videos of a particular task.

One aspect of understanding the procedure flow in the video is determining the step dependencies. Some earlier steps are prerequisites of later steps. For example, in \texttt{Carnitas Tacos With Cilantro Lime Sauce}, the sauce corresponding to steps 0 and 1 must be made, the meat corresponding to steps 2-7 must be cooked, and the tortilla corresponding to step 8 must be heated in order for the taco to be assembled in step 9. Therefore, steps 1, 7, and 8 are prerequisites of step 9, and this relationship is defined as {\bf sequential}. Meanwhile, steps 1, 7, and 8 deal with three parallel components of the taco, and switching their order will not affect the final dish. In other words, the step sequence 0,2,1,3,4,5,6,8,7,9 would also be a valid recipe resulting in the same dish. {\bf Parallel} relation is formally defined as different steps involving non-overlapping ingredients and utensils. Swapping the ordering of parallel steps will not affect the final dish. The sequential and parallel structure of the steps in a recipe can be characterized as a {\bf flow graph}, where directed edges connect sequential steps (represented as nodes), and the edge direction describes the execution order. All topological sorts of the flow graph will be valid recipes resulting in the same dish. A formal definition will be given in Section \ref{sec:method}. The flow graphs of the two recipes \texttt{Carnitas Tacos With Cilantro Lime Sauce} and \texttt{Carne Asada Tacos} are shown in Figure \ref{fig:example_flow} and \ref{fig:example_flow2} respectively. 

In this paper, we study the problem of predicting the flow graph given a procedural video instance and its step starting and ending timestamps. 
Some of the challenges associated with predicting flow graphs are: \textbf{First}, according to the definition of the parallel relation, the model needs to accurately recognize the involved ingredients and utensils, including distinguishing visually similar utensils and different ingredients. \textbf{Second}, cooking involves complex operations that transform ingredients both mechanically and chemically. Attributes such as shape and color can change drastically during this process. Consequently, the model needs to track the state change of the ingredients.

\begin{figure}[t]
    \centering
    \begin{subfigure}{1.0\linewidth}
        \centering
        \includegraphics[width=0.95\textwidth]{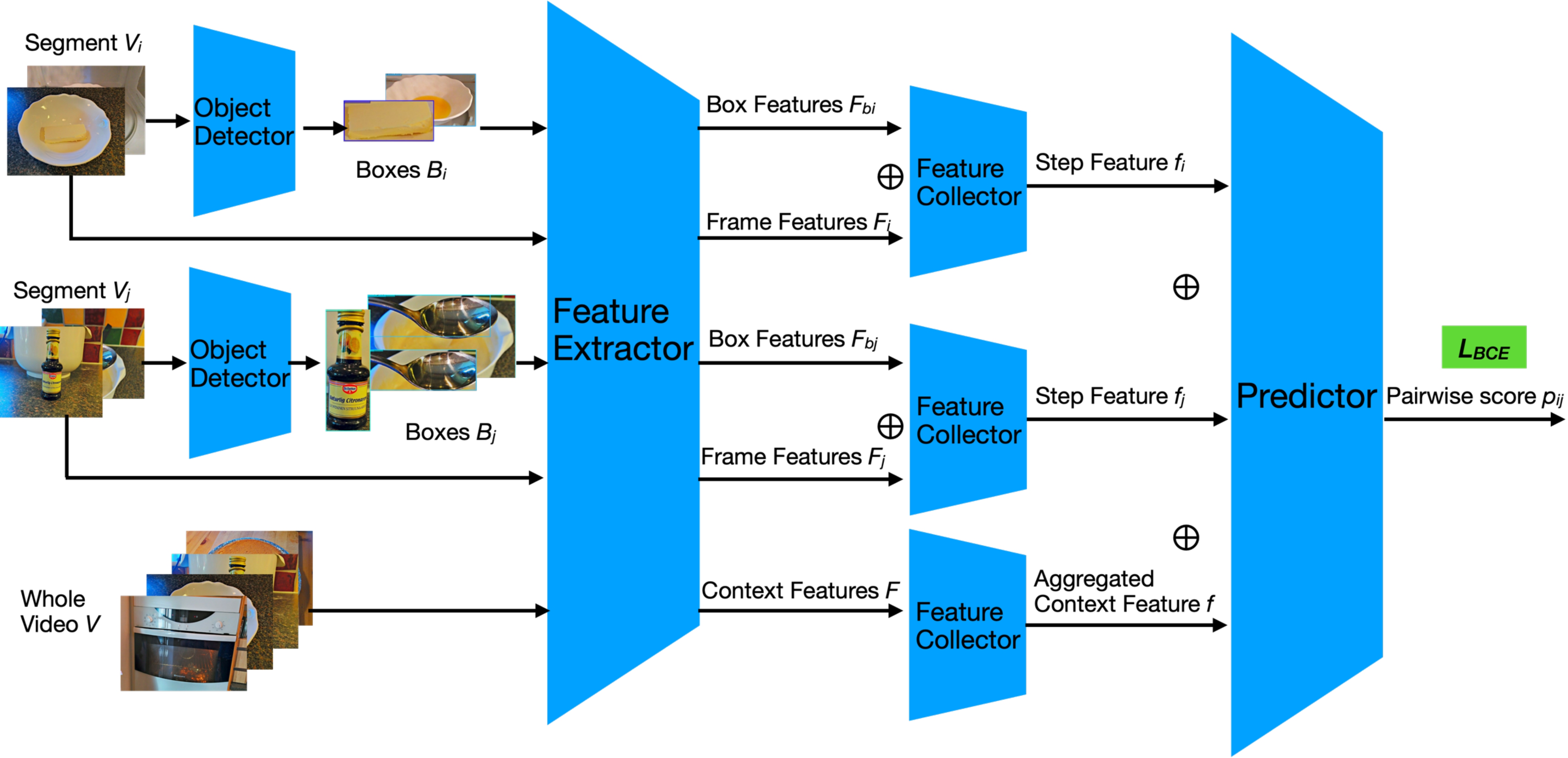}
        \vspace{-1\baselineskip}
        \caption{\centering \small Predicting pairwise relation scores from a pair of video step segments and the whole videos as context.}
    \end{subfigure}
    \begin{subfigure}{1.0\linewidth}
        \centering
        \vspace{0.3cm}
        \includegraphics[width=0.9\textwidth]{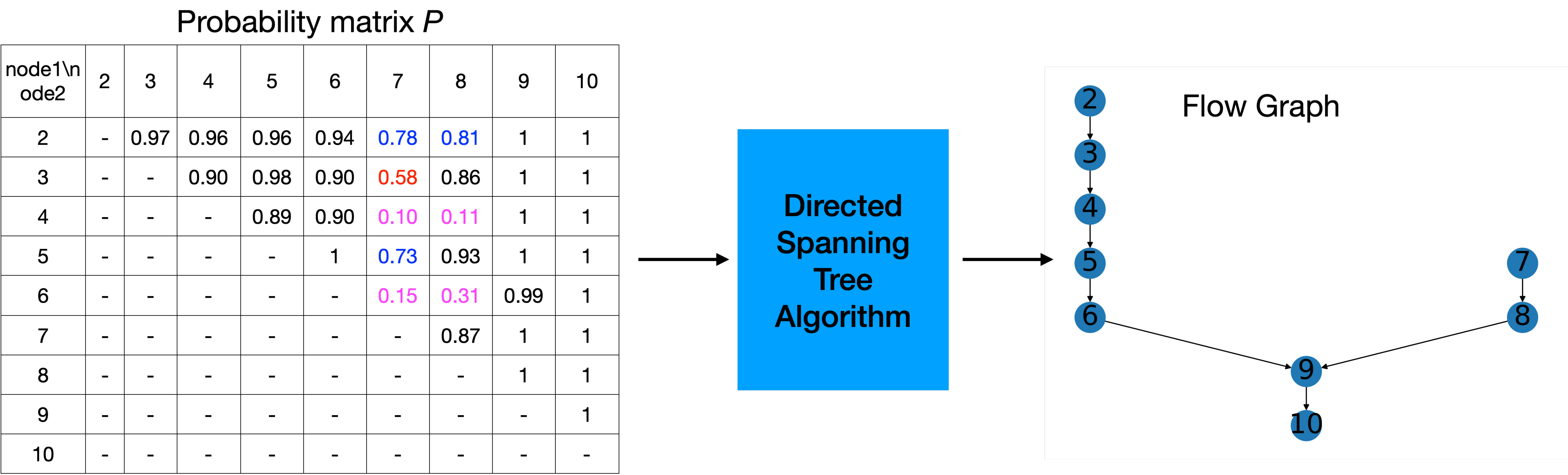}
        \vspace{-.5\baselineskip}
        \caption{\centering \small After calculating the edge probabilities for all pairs of steps in a video, we create the flow graph using a spanning tree algorithm for directed graphs.}
    \end{subfigure}
    \vspace{-2\baselineskip}
    \caption{Overview of our method. We first predict the edge probabilities for all step segment pairs then create the flow graph using a spanning tree algorithm from the probability matrix.}
    \vspace{-2\baselineskip}
    \label{fig:overview}
\end{figure}

To tackle these challenges, we propose the \texttt{Box2Flow} framework, as shown in Figure \ref{fig:overview}. We first calculate the edge probabilities for all step pairs in a video to get a probability matrix. Then, we create the flow graph from the matrix with a spanning tree algorithm for directed graphs. More specifically, to make our model focus on the ingredients and utensils involved in order to more accurately predict the step relations, we extract the object bounding boxes in the step segments. 
To tackle the second challenge, we include the whole video as context to monitor the state of each ingredient.
We experiment on the labeled MM-ReS\cite{pan2020multi} and the unlabeled YouCookII\cite{ZhXuCoAAAI18} datasets. Furthermore, we interpolated the missing frames in MM-ReS to improve the performance. In addition to traditional recall and precision metrics, we use maximal common subgraph\cite{bunke1998graph} for a more structural evaluation. Results show \texttt{Box2Flow} can effectively predict the flow graphs.

In summary, the contributions of this paper are:
\vspace{-.5\baselineskip}
\begin{itemize}
\item We study the less explored problem of generating the flow graph from a single procedural video instance.

\item We propose \texttt{Box2Flow} to solve the problem of predicting flow graphs from videos. Experiments show the framework effectively predicts the flow graphs.

\item We interpolate the missing frames in the MM-ReS dataset to extract flow graphs effectively. 
We also explore the utility of a learned flow graph predictor trained on MM-ReS to a zero-shot transfer task for the unlabeled YouCookII dataset.

\item We assess the accuracy of the predicted flow graphs to the ground truth graphs using the structural similarity metric, Maximal Common Subgraphs.
\end{itemize}

\vspace{-1\baselineskip}
\section{Related Work}
\vspace{-.5\baselineskip}
Our work relates to video graph representations, flow graph prediction, and their downstream applications. Previous work in computer vision has focused on various graph representations for videos. When characterizing tasks with multiple steps as flow graphs, previous work has focused on flow graph prediction and applications for text, sequences of images, and program codes. However, flow graph prediction from a single procedural video instance has been less explored.

{\noindent \bf Video graph representations. } One of the most studied graph representations is the scene graph. \cite{cong2021spatial,li2022dynamic}
generated video scene graphs by predicting the scene graphs at each frame. Although the temporal properties were considered during prediction, the outputs for each frame remain unconnected. \cite{holm2023dynamic} added temporal edges between the same node in neighboring frames, which is also called spatio-temporal graph. \cite{wu2021scenegraphfusion} built a panoptic 3D scene from RGB-D videos by exploiting the actual spatial relations between neighboring scenes. For representing videos as various types of graphs, \cite{hussein2019videograph,schiappa2022svgraph} 
learned a semantic graph from the entire instructional video where the nodes learn semantic concepts and the edges are calculated from the node features. \cite{jang2023multimodal,mao2023action}
learned {\bf general} flow graphs from {\bf multiple} videos for {\bf each task} and \cite{zhou2023procedure} built a large flow graph together for all tasks. In these non-instance-based works, if two steps are performed in different orderings in different videos, they are treated as parallel. However, the parallel relationships might not be fully covered due to the limited number of videos in the dataset. On the other hand, the relationships between the steps are intrinsic to the steps themselves through the involved objects. To address the issues caused by task-based methods that learn a single graph from multiple videos, in our work, we focus on an instance-based method that learn flow graphs specific to each available video. This approach allows us to develop more precise and detailed video-specific representations.

Regarding graph downstream applications, \cite{ost2021neural,cong2023ssgvs} used scene graphs for video rendering and synthesis, respectively. \cite{rodriguez2021dori,ou2022object} used spatio-temporal graphs for temporal moment localization given language queries and action recognition, respectively. Specifically, \cite{rodriguez2021dori} included linguistic and visual nodes in their graph. \cite{xu2020g} formulated video snippets as graph nodes and snippet correlations as edges for action detection. \cite{bar2020compositional} used task graphs where nodes correspond to objects and edges correspond to actions for video synthesis. \cite{huang2019neural} represented videos as conjugate task graphs where the nodes are actions, and the edges are states for a single-shot action plan. \cite{huang2020improving} used graph representations for action segmentation where the nodes are segments, and the edges represent neighboring segment relations. \cite{tu2023relation} used graphs for video captioning where the nodes include both whole video features and word features. 

{\noindent \bf Flow graph prediction and downstream applications.} \cite{mori2014flow,yamakata2020english} created the fine-grained flow graphs from Japaneses and English recipe texts respectively where each node is a named entity. \cite{pan2020multi,pan2020hybrid,zhang2022miais}
created recipe step flow graphs from text and images. Specifically, \cite{zhang2022miais} used Japanese language text.

In terms of applications employing flow graphs, \cite{luo2021operation} used flow graphs from 5G communication base station product manual texts for error detection and correction. \cite{bobrovnikova2022technique} used program control flow graphs for malware detection. \cite{dvornik2022graph2vid} used {\bf general} task flow graphs for video grounding, which were created from web text. Specifically, only {\bf one} flow graph is created for all videos of the same category. \cite{nishimura2020structure} used the help of flow graphs from text to train a captioning model where the inputs include a list of ingredients and a sequence of images where each image is considered as a single step.

In summary, instance-based flow graph prediction 
from a single video has been less explored. Compared with text, predicting flow graphs from visual inputs are more challenging. The involved objects might not be salient in images and videos. The view points might also change and the cooking process will drastically change the visual appearances of the ingredients, making them difficult to track therefore challenging to create the flow. Meanwhile, long videos contain more information than a sequence of a few frames (less than 100 images) and, subsequently, are more challenging. \cite{pan2020multi,pan2020hybrid} are image-based methods which average all image features in a single step while we study video-instance inputs. We explicitly model the input images or video clips as sequences to capture the action information. Furthermore, predicting a flow graph for each video can preserve the specific steps in the recipe that might not be covered by the general task flow graph, e.g., A general flow graph of the task \texttt{making coffee} might miss some steps specific to certain videos including \texttt{add milk foam} and \texttt{add syrup} which gives unique flavoring to the recipe. These steps could be included when predicting the flow graph from one video instance. In this paper, we predict a flow graph for each video instance by predicting pairwise edge probabilities from both frame-level and object features, then convert the probability matrix to a flow graph with a spanning tree algorithm.

\vspace{-1\baselineskip}
\section{Method}
\label{sec:method}
\vspace{-.5\baselineskip}
\subsection{Flow Graph Definition}
\vspace{-.5\baselineskip}
\label{sec:def}
A flow graph $F=(S,E)$ is a directed acyclic graph where each node is a step. Let the set of nodes $S=\{S_1,S_2,\ldots,S_i,\ldots,S_n\}$, where $S_i$ is the $i$-th step in the recipe. A directed edge $(S_i,S_j)$ exists between if and only if the following rules hold:
\vspace{-.5\baselineskip}
\begin{enumerate}
    \item $i<j$, and
    \item $S_i$ and $S_j$ are sequential, and
    \item if $j>i+1$, there is no such $k$ where $i<k<j$, such that both step pairs $(S_i, S_k)$ and ($S_k, S_j$) are sequential.
\end{enumerate}
Rule 1 determines the graph's flow where the later nodes are descendants. Rule 2 considers the sequential relation as edges, and rule 3 does not allow skip edges. 

In other words, an edge connects a step with its direct consequence. If $S_i$ and $S_j$ are sequential but indirect (e.g., steps 2 and 9 in \texttt{Carnitas Tacos With Cilantro Lime Sauce Recipe}, as shown in Figure \ref{fig:recipe_intro1}, \ref{fig:example_flow}), there exists a path with length at least two between $S_i$ and $S_j$ and vice versa.
\vspace{-1\baselineskip}
\subsection{Pairwise Edge Probability}
\vspace{-.5\baselineskip}
Given a video $V$ and the start and end frames for each step $T=\{(s_i,e_i)|1\leq i\leq n\}$, our goal is to predict the flow graph $F$ or the set of edges $E\subseteq \{(S_i,S_j)|1\leq i<j\leq n\}$. In addition, we denote the $i$-th video segment $V[s_i:e_i]$ as $V_i$.

To predict the pairwise edge probability, we first extract the object bounding boxes with an object detector:
\vspace{-1\baselineskip}
\begin{equation}
    B_i=F_{od}(V_i)
\vspace{-.5\baselineskip}
\end{equation}
where $B_i=\{(x_{{min}_{kt}},y_{{min}_{kt}},x_{{max}_{kt}},y_{{max}_{kt}})\}. (x_{{min}_{kt}},y_{{min}_{kt}})$ is the top-left corner and $(x_{{max}_{kt}},y_{{max}_{kt}})$ is the bottom-right corner of the $k$-th box in the $t$-th frame of the segment. The frame patches defined by $B_i$ are denoted as $V_i[B_i]$.

Next, we extract the object features with a video encoder:
\vspace{-.5\baselineskip}
\begin{equation}
    F_{b_i}=F_{fe}(V_i[B_i])
\vspace{-.5\baselineskip}
\end{equation}
where $F_{b_i}\in \Real^{K_i\times d}$. $K_i=\sum_{t=1}^{e_i-s_i+1}k_t$ is the total number of bounding boxes in the segment and $d$ is the output feature dimension.

Similarly, the frame features of $V_i$ can be extracted as 
\vspace{-.5\baselineskip}
\begin{equation}
    F_i=F_{fe}(V_i)
\vspace{-.5\baselineskip}
\end{equation}
where the bounding boxes can be treated as (1,1,$W$,$H$) for all frames. $W$ is the frame width and $H$ is the frame height. $F_i\in \Real^{(e_i-s_i+1))\times d}$.

As the step relations given only two video segments can be ambiguous, we also include the whole video feature $F$ as context, which consists of all frame-level features stacked together:
\vspace{-.5\baselineskip}
\begin{equation}
F=stack(F_1,F_2,\ldots,F_i,\ldots,F_n)
\vspace{-.5\baselineskip}
\end{equation}
$F\in \Real^{N\times d}$ where $N=\sum_{i=1}^n(e_i-s_i+1)$ is the total number of non-background frames related to the task.

Then, we aggregate the frame and box features for each segment using BERT with adapters\cite{houlsby2019parameter}. The features are first projected to BERT input embedding dimension through the same linear layer:
\vspace{-.5\baselineskip}
\begin{gather}
    F_{e_i}=\tanh\left( F_{fc}(F_i)\right) \\
    F_{e_{b_i}}=\tanh \left( F_{fc}(F_{b_i})\right)\\
    F_e=\tanh \left( F_{fc}(F)\right)
\end{gather}

The frame and the box embeddings are stacked together with the BERT [CLS] embedding and fed through the transformer encoder to extract the step features. The position IDs for the $k_t$ boxes and the frame embeddings in the $t$-th frame are all $t$, and the [CLS] embedding has position ID 0. The output of [CLS] representation is taken as the aggregated feature. For the context feature, only the frame feature $F$ is used.
\vspace{-.5\baselineskip}
\begin{gather}
f_i=F_{bert_{step}}\left(stack(F_{cls},F_{e_i},F_{e_{b_i}})\right)    \label{eq:feature_vi}\\
    f=F_{bert_{ctx}}\left(stack(F_{cls},F_e)\right)
    \label{eq:feature_v}
\vspace{-.5\baselineskip}
\end{gather}
where $F_{bert_{step}}, F_{bert_{ctx}}$ are two different BERT adapters for step features and context features respectively. $F_{cls}$ is BERT [CLS] embedding. $f_i,f\in \Real^{d_{bert}}$ are 1-D vectors with BERT output dimension.

Finally, two-step features $f_i,f_j, (i<j)$ and the context feature $f$ are concatenated and fed through an MLP to predict the pairwise sequential probability:
\vspace{-.5\baselineskip}
\begin{equation}
    p_{ij}=\sigma\left(F_{mlp}(f_i\oplus f_j\oplus f)\right)
    \label{eq:prob}
\end{equation}
where $\sigma(\cdot)$ is the Sigmoid function and $\oplus$ stands for concatenation.

During training, we use the weighted binary cross-entropy loss:
\vspace{-.5\baselineskip}
\begin{equation}
    L=-w_s\sum_{1\leq i<j\leq n}[w_py_{ij}\log p_{ij}+(1-y_{ij})\log (1-p_{ij})],
\vspace{-.5\baselineskip}
\end{equation}
where $w_s$ is the video sample weight, $w_p$ is the positive weight for unbalanced label distribution. $y_{ij}=1$ for sequential relation, both direct and indirect, and $y_{ij}=0$ for parallel relation.

{\bf Multi-modality.} Our framework can be easily extended to prediction with only text or both video and text modalities. When only the recipe text is available, we have $R=R_1\oplus R_2\oplus\ldots \oplus R_i\oplus\ldots \oplus R_n $, where $R_i$ is the text for the $i$-th step. Then, the text tokens are directly fed to the BERT adapters to get the features for the MLP module in Equation \ref{eq:prob}:
\vspace{-.5\baselineskip}
\begin{gather}
    f_{text_i}=F_{bert_{text}}(R_i)\label{eq:feature_ti}\\
    f_{text}=F_{bert_{text}}(R)
    \label{eq:feature_t}
\vspace{-.5\baselineskip}
\end{gather}
The text-described step and context features share the same adapter. The sequential probability is calculated as follows:
\vspace{-.5\baselineskip}
\begin{equation}
    p_{ij}=\sigma\left(F_{mlp}(f_{text_i}\oplus f_{text_j}\oplus f_{text})\right)
    \label{eq:text_prob}
\vspace{-.5\baselineskip}
\end{equation}
When both video and text are available, after the features $f_i,f,f_{text_i}, f_{text}$ are extracted as Equation \ref{eq:feature_vi}, \ref{eq:feature_v}, \ref{eq:feature_ti}, \ref{eq:feature_t}, the probability is calculated as:
\vspace{-.5\baselineskip}
\begin{equation}
    p_{ij}=\sigma\left(F_{mlp}(f_i\oplus f_j\oplus f \oplus f_{text_i}\oplus f_{text_j}\oplus f_{text})\right)
\vspace{-.5\baselineskip}
\end{equation}
Three different adapters are involved: video step, video context and text.

\subsection{Graph Construction}

We construct the flow graph after all the pairwise scores $P=\{p_{ij}|1\leq i<j\leq n\}$ have been calculated. Since most of the flow graphs in the real world are trees with at most one descendent for each node, we focus on constructing trees. Because of rule 3 in Section \ref{sec:def}, if the flow graph is a tree and there is an edge between $(S_i,S_j)$, $S_j$ has to be the earliest step such that $S_i$ and $S_j$ are sequential.

\begin{wrapfigure}{R}{0.5\textwidth}
    \begin{minipage}{0.5\textwidth}
    \vspace{-3.5\baselineskip}
      \begin{algorithm}[H]
        \caption{Flow graph from probability matrix}
        \begin{algorithmic}
    \State \textbf{Inputs:} probability matrix $P$, number of steps $n$
    \State \textbf{Output:} edge set $E$
    \State $E \leftarrow \phi$
    \State $E_{can}\leftarrow \{(S_i,S_j)|p_{ij}>0.5\}$
    \For{i = 1 to n}
    \State $I\leftarrow\{j|(S_i,S_j)\in E_{can}\}$
    \If{$I\neq \phi$}
    \State $j\leftarrow argmin_{j>i}(j\in I)$
    \State $E\leftarrow E\cup\{(S_i,S_j)\}$
    \EndIf
    \EndFor
    \State \textbf{Return} E
    \vspace{-.2\baselineskip}
    \end{algorithmic}
    \label{alg:flow}
    \end{algorithm}
    \end{minipage}
\end{wrapfigure}

Therefore, we first select the edges according to the standard probability threshold 0.5 to get a set of candidate edges $E_{can}=\{(S_i,S_j)|p_{ij}>0.5\}$. Then, for each step $i$, we select the earliest step $j$ such that  $(S_i,S_j)\in E_{can}$ to form the flow graph, as in Algorithm \ref{alg:flow}.

    
\section{Experiments}
\subsection{Datasets and Metrics}

We use two datasets, labeled MM-ReS\cite{pan2020multi} and unlabeled YouCookII\cite{ZhXuCoAAAI18}.\\

{\bf MM-ReS} consists of recipe texts, step images and annotated flow graphs. The original dataset includes 9850 recipes collected from the Internet. 
Since we focus on predicting tree graphs, we only use the 8370 recipes with tree annotations and randomly split into training, validation and test sets, with 6696, 837, 837 each. 64.5k steps are annotated in flow graphs. The dataset includes 131k images, with 2.8 images/step on average for steps with images. The images can be treated as short video clips for each step. Meanwhile, 18\% of the steps do not have images and need to be removed, zero-padded or interpolated.

{\bf YouCookII} consists of cooking videos from YouTube, with step starting and ending time annotations and step text descriptions rephrased by annotators but without annotated flow graphs. 
We use 1181 training videos and 414 validation videos, which are still available on Youtube.
There are 7.7 steps/video on average. For evaluation, we manually annotated the flow graphs of 39 videos in the training set and 63 videos in the validation set. 97 of the annotations are trees. During training, we combined the manual annotation and predictions from a text model trained on MM-ReS as labels.

{\bf Metrics.} Following \cite{pan2020multi}, we report edge-level recall $R_e$, precision $P_e$, F1 and recipe-level recall $R_r$, precision $P_r$, F1, which are calculated as the follows:

Suppose the ground truth edge set of the $i$-th video in the dataset is $E_i$, the predicted edge set is $\hat{E_i}$, $|\cdot|$ denotes set cardinality and there are $M$ videos in the dataset,
\vspace{-.5\baselineskip}
\begin{align}
    R_e=\frac{\sum_{i=1}^M|E_i\cap \hat{E_i}|}{\sum_{i=1}^M|E_i|},\  
    P_e=\frac{\sum_{i=1}^M|E_i\cap \hat{E_i}|}{\sum_{i=1}^M|\hat{E_i}|}
    \vspace{-.5\baselineskip}
\end{align}
The edge-level $F_1$ is the harmonic average between $R_e$ and $P_e$.
Define $R_i, P_i$ as the precision and recall for each recipe, calculated as 
\vspace{-.5\baselineskip}
\begin{align}
    R_i=\frac{|E_i\cap \hat{E_i}|}{|E_i|},\  
    P_i=\frac{|E_i\cap \hat{E_i}|}{|\hat{E_i}|}
    \vspace{-.5\baselineskip}
\end{align}
Define $F1_i$ as the harmonic average between $R_i$ and $P_i$. Then $R_r, P_r, F1_r$ are calculated as $\frac{1}{M}\sum_{i=1}^MR_i, \frac{1}{M}\sum_{i=1}^MP_i, \frac{1}{M}\sum_{i=1}^MF1_i$ respectively.

\begin{figure}[t]
    \centering
    \begin{subfigure}[t]{0.45\linewidth}
        \centering
        \includegraphics[width=1.0\textwidth]{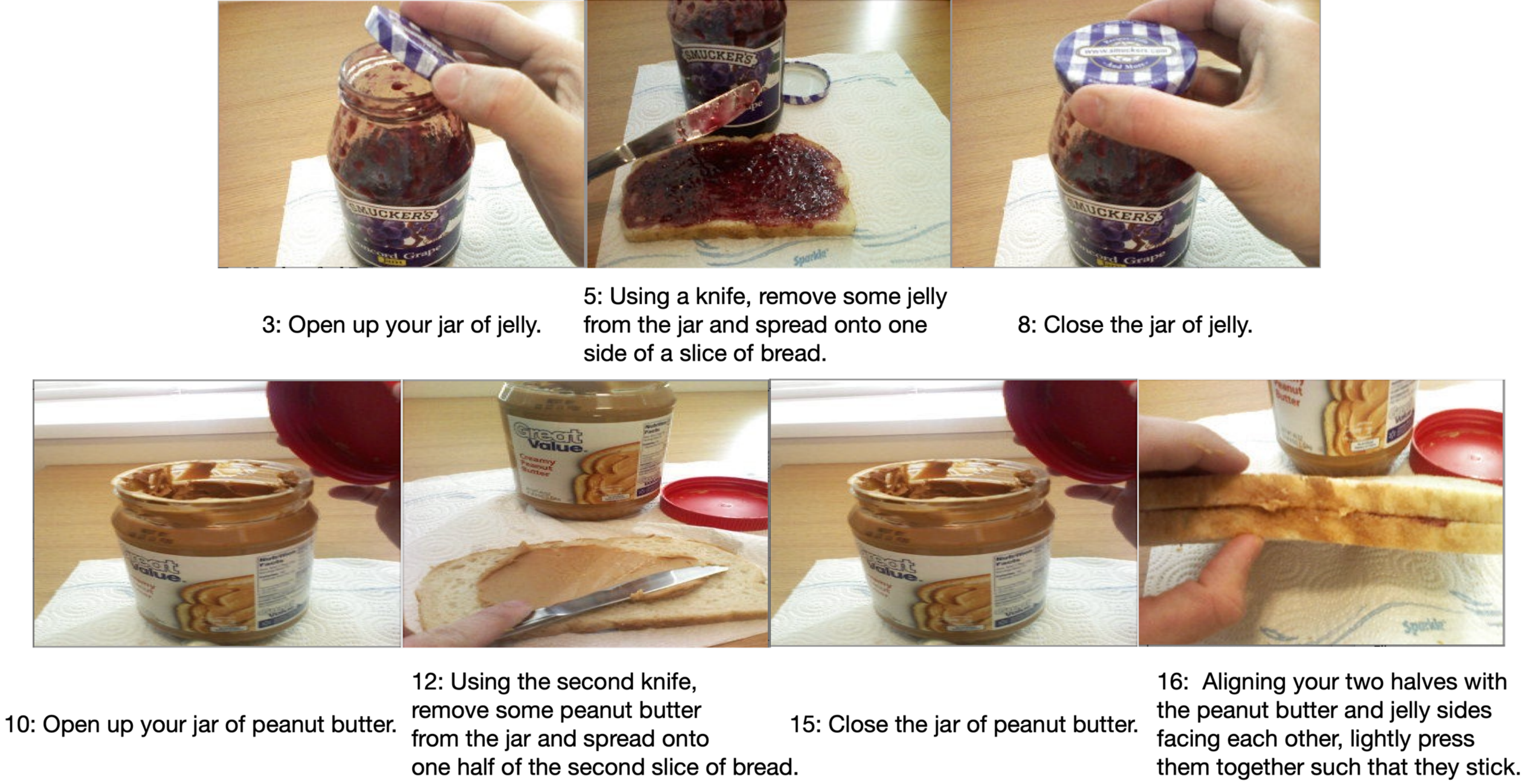}
    \vspace{-1\baselineskip}
        \caption{\centering \small Peanut Butter and Jelly Sandwich. }
    \end{subfigure}
    \begin{subfigure}[t]{0.25\linewidth}
        \centering
        \includegraphics[width=1.0\textwidth]{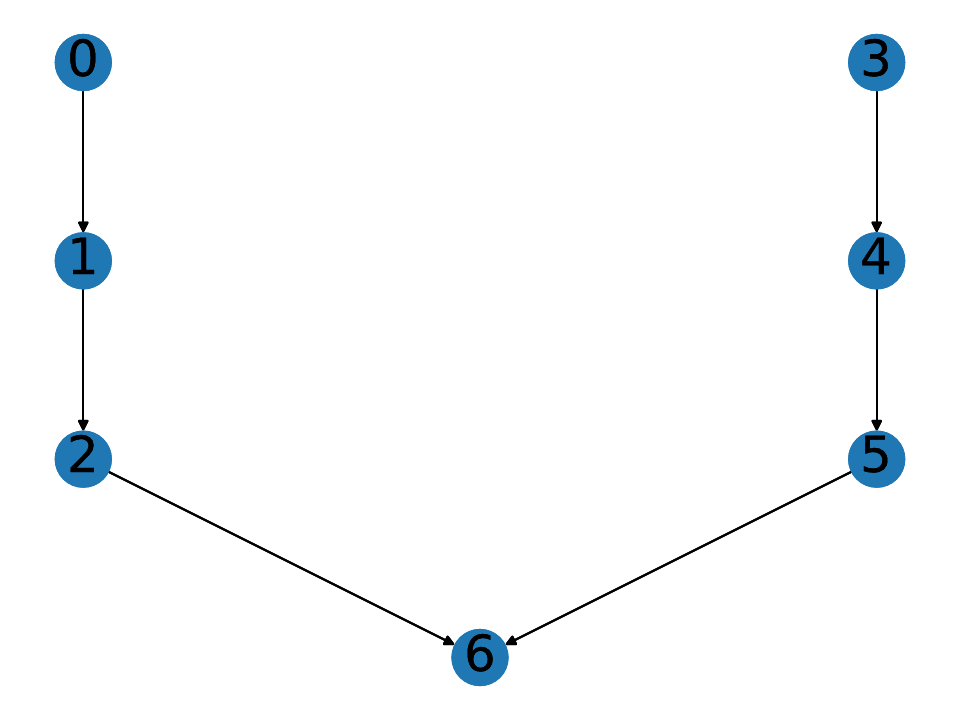}
        \caption{\centering \small Ground truth flow graph. }
    \end{subfigure}
    \begin{subfigure}[t]{0.25\linewidth}
        \centering
        \includegraphics[width=1.0\textwidth]{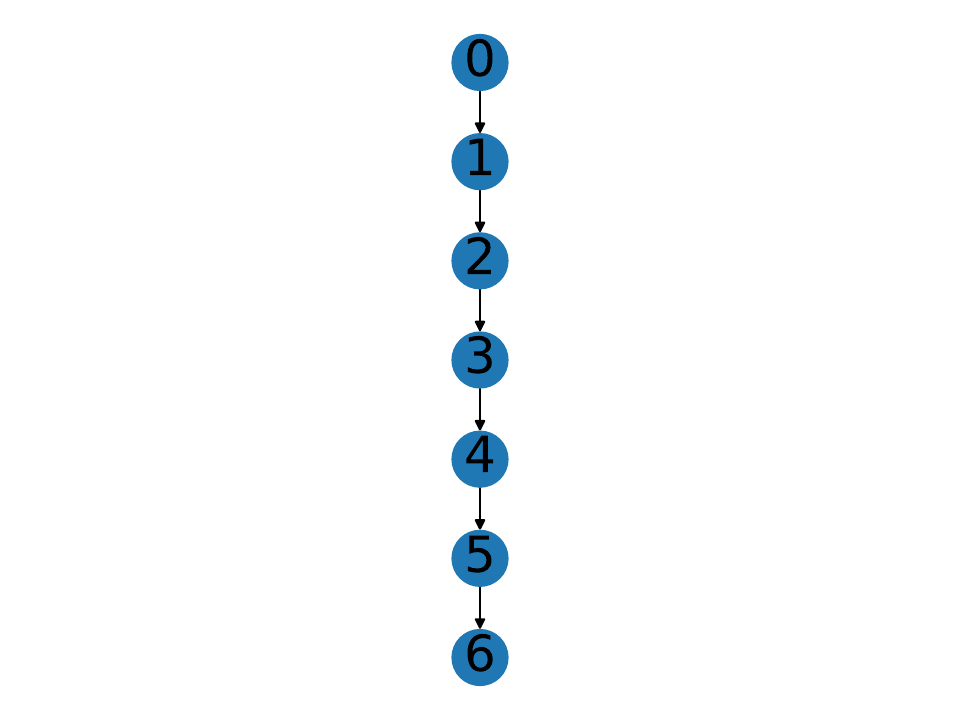}
        \caption{\centering \small Predicted flow graph.}
    \end{subfigure}
    \vspace{-0.8\baselineskip}
    \caption{An example of ground truth and predicted flow graphs where the recall and precision are high but very different structurally. The recipe is \texttt{Peanut Butter and Jelly Sandwich} from MM-Res.}
    \vspace{-1\baselineskip}
    \label{fig:sim}
\end{figure}

However, these metrics might not accurately reflect the structural similarity between the predicted and the ground truth graph, as shown in Figure \ref{fig:sim}, taken from \texttt{Peanut-Butter-and-Jelly-Sandwich-1} recipe in MM-ReS. The ground truth shows two branches merging, while the predicted is a chain. Therefore, the structures are very different. However, the only different edge is (2,6) in the ground truth and (2,3) in the predicted. Recall, precision, and F1 are all as high as 83\% in this case.
As a result, we also include a structural similarity metric maximal common subgraph (MCS)\cite{bunke1998graph}. Define $cc$ as the number of nodes in the connected component of $E\cap\hat{E}$ with the maximum size and $n$ is the number of nodes in $E$, $MCS=cc/n$. In the example, the maximal common subgraph is \{(3,4),(4,5),(5,6)\} with 4 nodes and mcs=4/7=57\%.

\vspace{-1\baselineskip}
\subsection{Compared Methods}
\vspace{-.5\baselineskip}
We investigate methods using different modalities, including video-only, text-only, and video+text. \footnote{Our results are not directly comparable with \cite{pan2020multi,pan2020hybrid} because of different evaluation subsets and code not available.}Specifically since only a few annotations are available for YouCookII, we directly transfer a pre-trained model on MM-ReS for text-only to show zero-shot ability.

We include video captioning as a baseline for video-only methods. Captions are first generated for videos then flow graphs are created from the captions using Equation \ref{eq:text_prob}. We also manually annotated some examples from generated captions. The details are in our supplement. 
We compare with the baseline video captioning methods {\bf MART}\cite{lei2020mart} and {\bf VLTinT}\cite{yamazaki2023vltint}.

To show the effects of bounding boxes, we compare \texttt{Box2Flow} with its variance using only frame features $F_{e_i}$ in Equation \ref{eq:feature_vi} but not box features $F_{e_{b_i}}$. The variance is denoted by "f". To remove the effects of more parameters introduced by two adapters, we also include models using the same adapter for both context and step features, denoted by "1". Specifically, for video+text methods on YouCookII, all methods are trained with one adapter.

For MM-ReS, we fix bottom-up attention\cite{Anderson2017up-down} for feature extraction and compare two different object detectors, 
Detectron2\cite{wu2019detectron2} 
pre-trained on COCO\cite{lin2014microsoft} and SAM\cite{kirillov2023segment} masks, denoted by "C" and "S" respectively. As 18\% of the steps do not have images in MM-ReS, we also study the effect of interpolating the missing images using instruct-pix2pix\cite{brooks2023instructpix2pix} for image+text models, denoted by "i". Otherwise, we zero-pad the image features for image+text methods and directly remove these nodes for image-only methods.

For YouCookII, we compare two different frame feature extractors, 
Densecap \cite{zhou2018end} and SlowFast\cite{feichtenhofer2019slowfast}, 
denoted by "D" and "SF" respectively. Specifically, only SlowFast can include bounding boxes for feature extraction.
We fix Detectron2 as the object detector.

We also include a naive {\bf chain} baseline which only requires the number of steps: $E=\{(S_1,S_2), (S_2,S_3),\ldots,(S_i,S_i+1),\ldots,(S_{n-1},S_n)\}$.

The implementation details are in our Supplement.
\vspace{-1\baselineskip}
\begin{figure}[h]
    \centering
    \begin{subfigure}[t]{0.45\linewidth}
        \centering
        \includegraphics[width=1.0\textwidth]{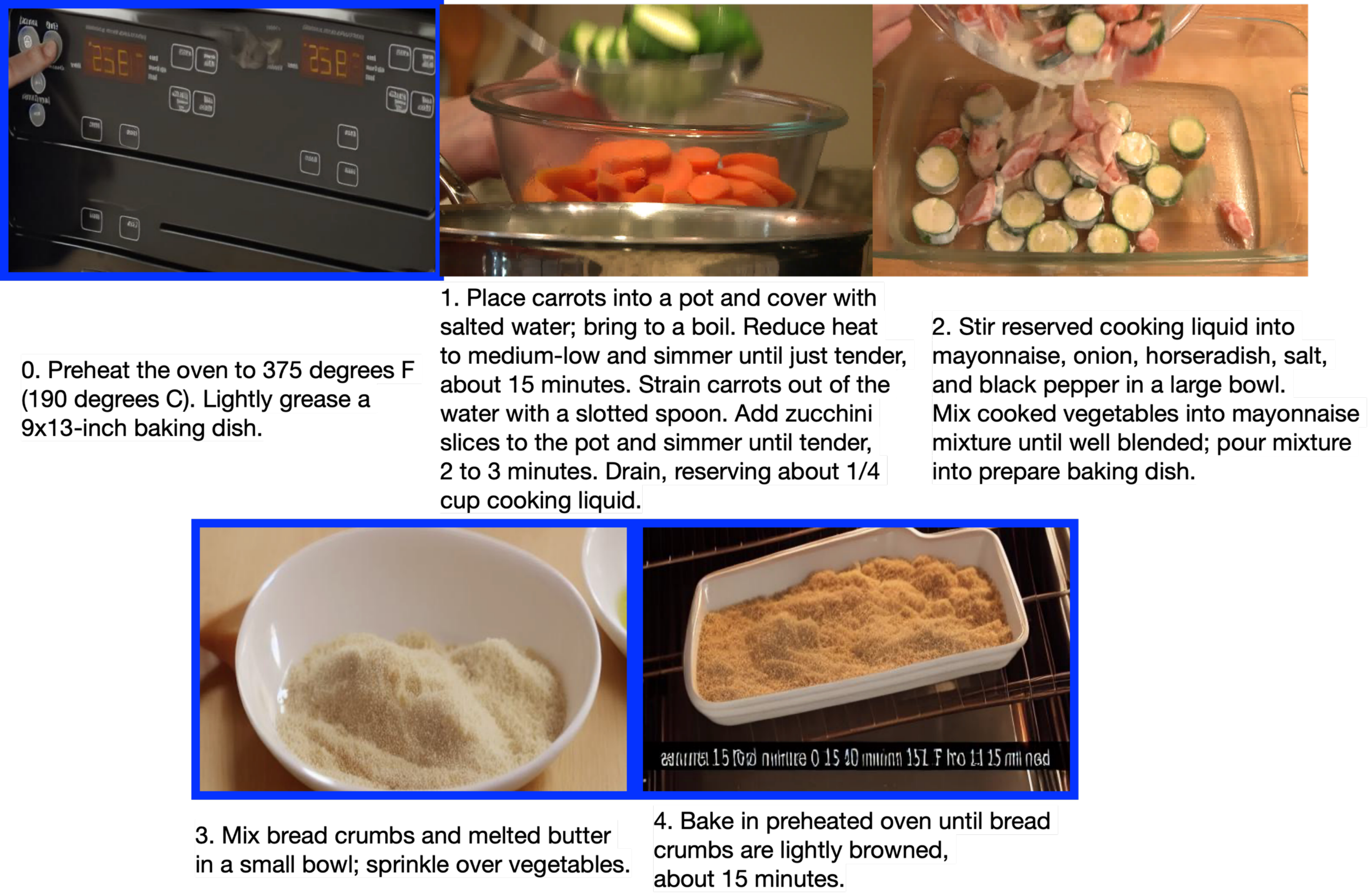}
    \vspace{-1.5\baselineskip}
        \caption{\centering \small A vegetable bake recipe. }
    \end{subfigure}
    \begin{subfigure}[t]{0.3\linewidth}
        \centering
        \includegraphics[width=1.0\textwidth]{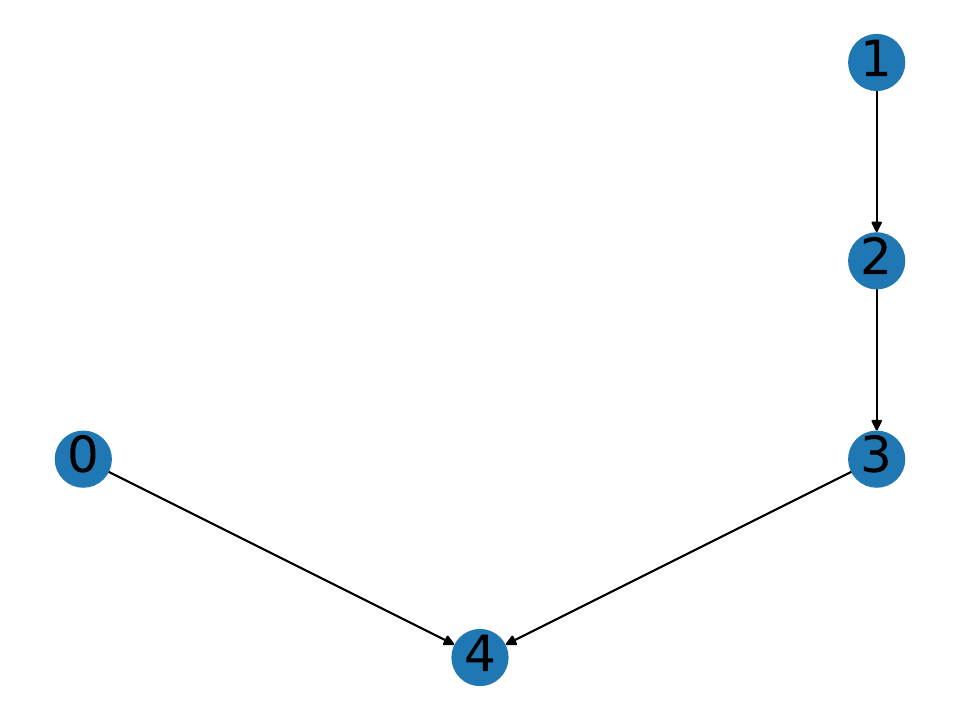}
    \vspace{-2\baselineskip}\caption{\centering \small Ground truth flow graph. }
    \end{subfigure}
    \begin{subfigure}[t]{0.3\linewidth}
        \centering
        \includegraphics[width=1.0\textwidth]{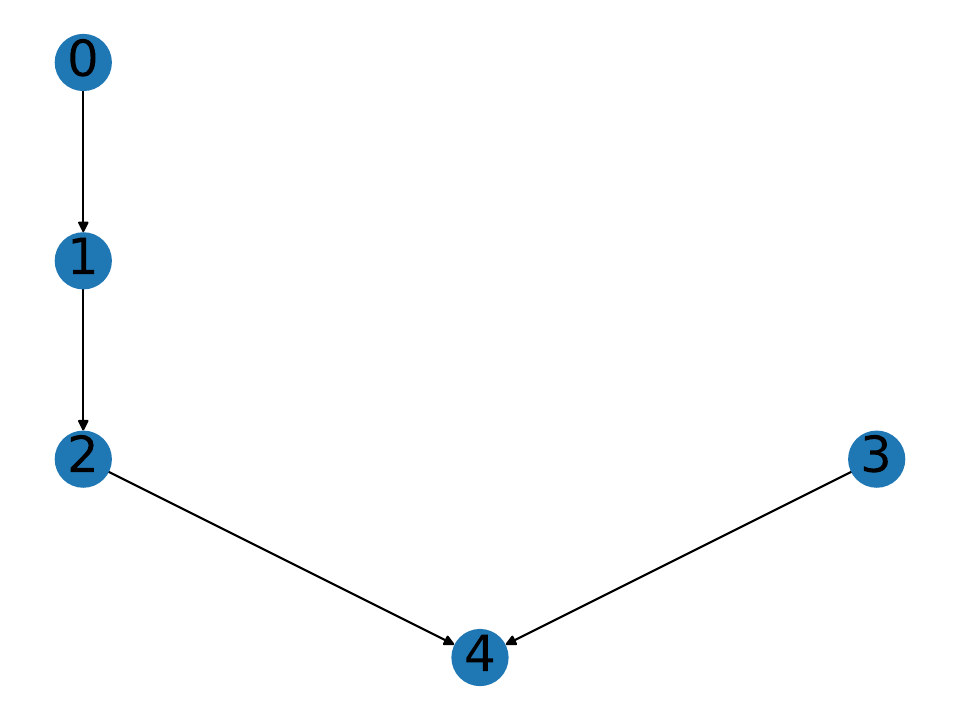}
    \vspace{-2\baselineskip}
        \caption{\centering \small Predicted flow graph from text model.}
    \end{subfigure}
    \begin{subfigure}[t]{0.3\linewidth}
        \centering
        \includegraphics[width=1.0\textwidth]{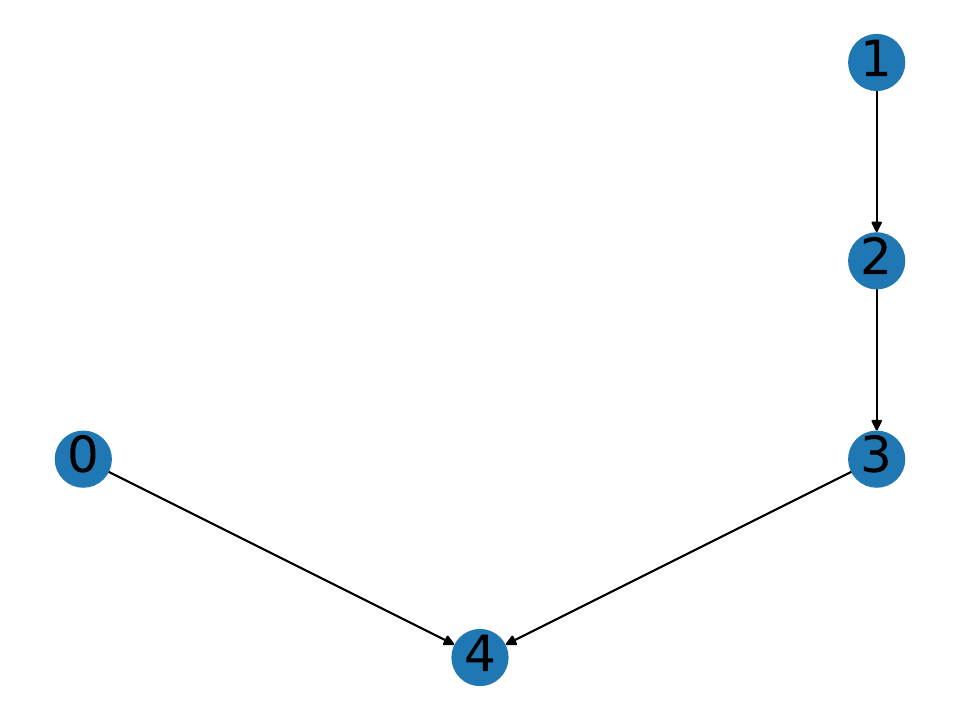}
    \vspace{-2\baselineskip}
        \caption{\centering \small Predicted flow graph from image+text model.}
    \end{subfigure}
    \vspace{-1\baselineskip}
    \caption{An example from MM-ReS. The text-only model did not predict the graph correctly, while the image+text model did. The interpolated images are marked in \textcolor{blue}{blue}.}
    \label{fig:mmres}
\end{figure}
\vspace{-1\baselineskip}

\begin{figure}[!]
    \centering
    \begin{subfigure}[t]{0.55\linewidth}
        \centering
        \includegraphics[width=1.0\textwidth]{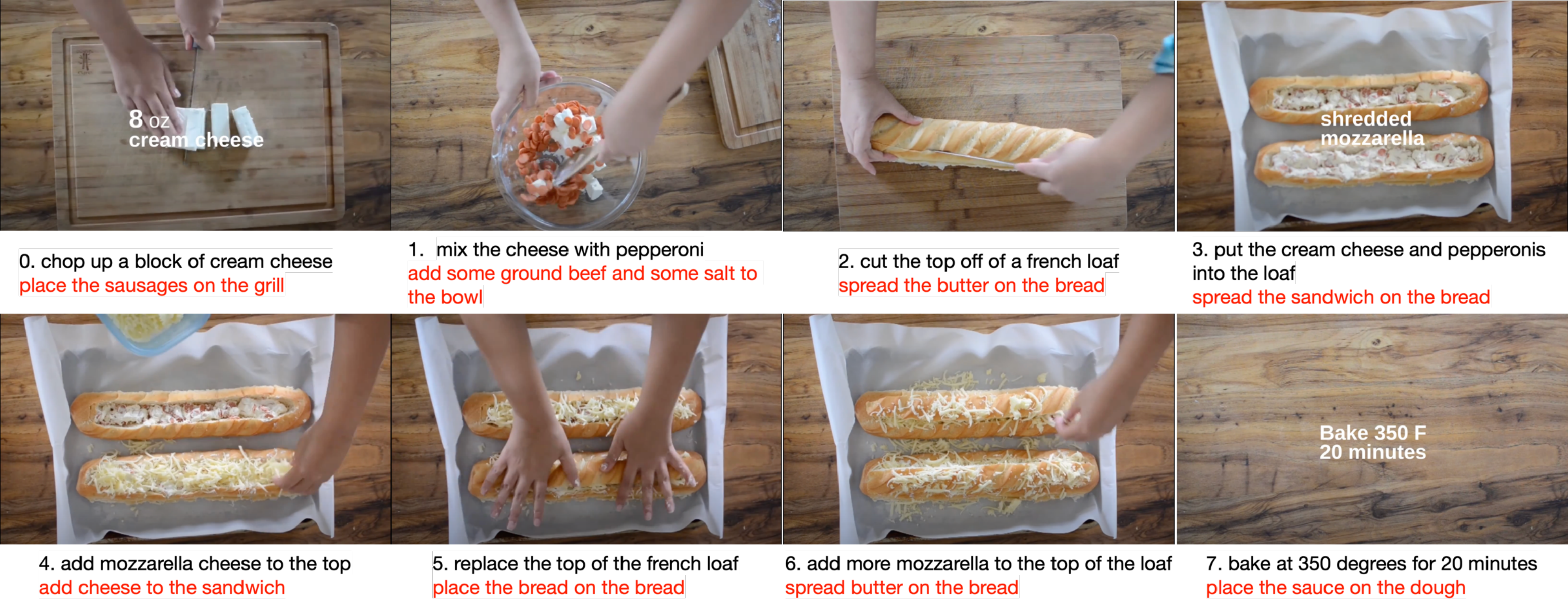}
    \vspace{-1.5\baselineskip}
        \caption{\centering \small Yummy Pepperoni Pizza Bread }
    \end{subfigure}
    \begin{subfigure}[t]{0.3\linewidth}
        \centering
        \includegraphics[width=0.95\textwidth]{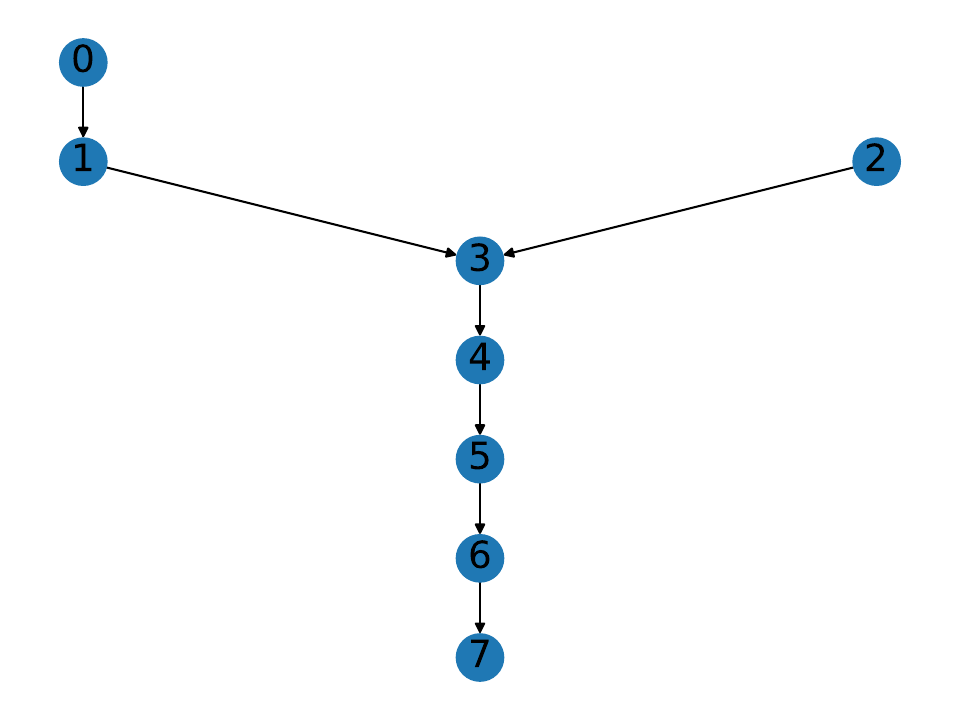}
    \vspace{-1\baselineskip}
        \caption{\centering \small Ground truth flow graph. }
    \end{subfigure}
    \begin{subfigure}[t]{0.24\linewidth}
        \centering
        \includegraphics[width=1.0\textwidth]{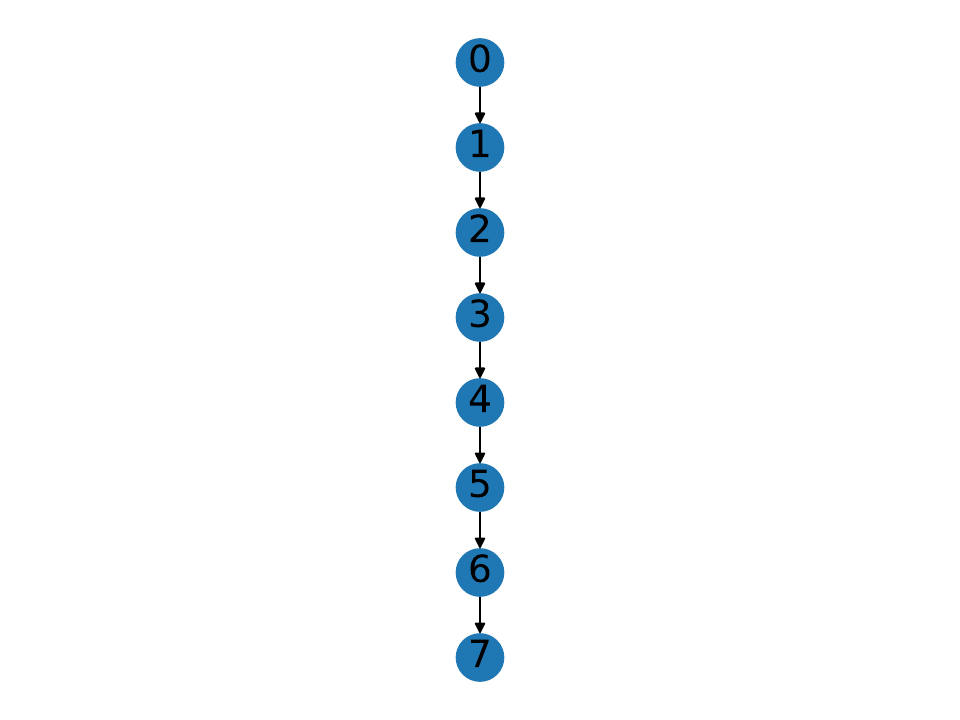}
    \vspace{-2\baselineskip}
        \caption{\centering \small Predicted flow graph from video captioning.}
    \end{subfigure}
    \begin{subfigure}[t]{0.24\linewidth}
        \centering
        \includegraphics[width=1.0\textwidth]{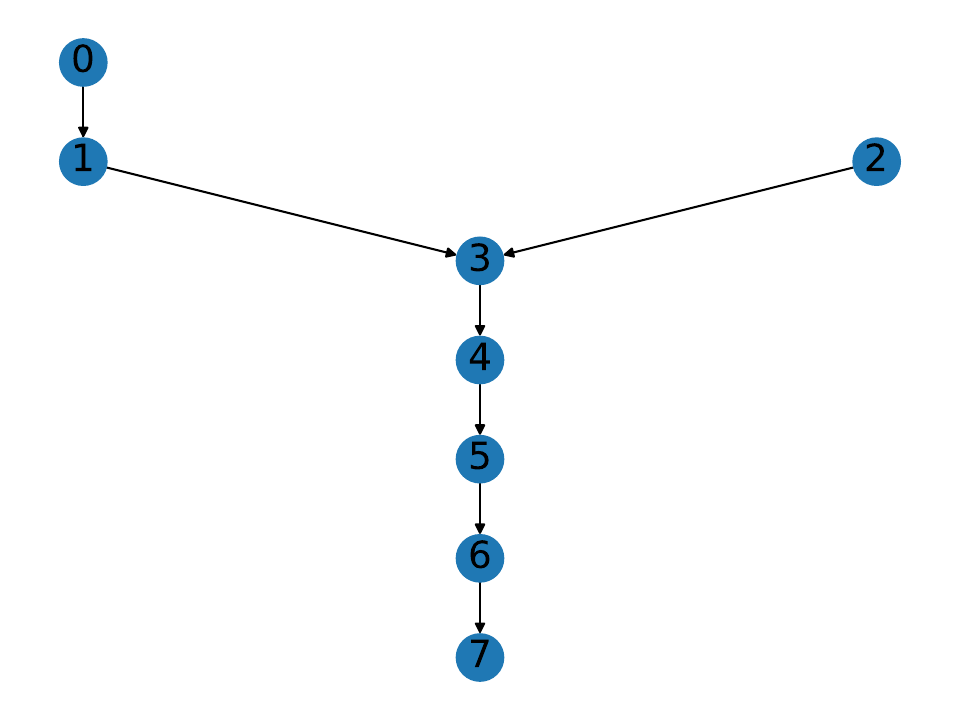}
    \vspace{-2\baselineskip}
        \caption{\centering \small Predicted flow graph from video model.}
    \end{subfigure}
    \begin{subfigure}[t]{0.24\linewidth}
        \centering
        \includegraphics[width=0.95\textwidth]{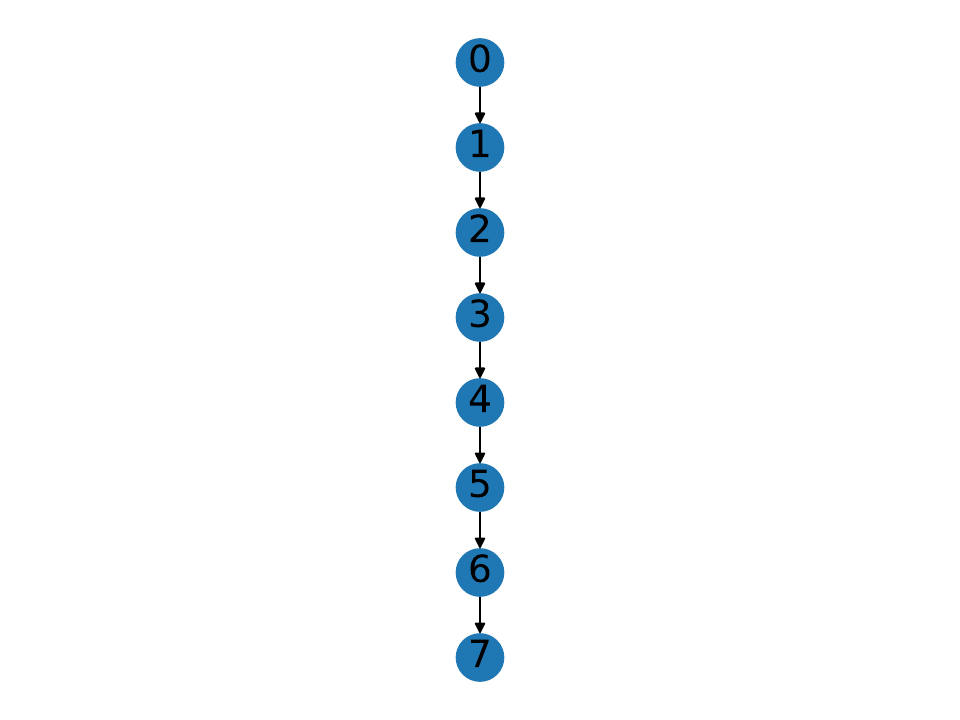}
    \vspace{-1\baselineskip}
        \caption{\centering \small Predicted flow graph from text model.}
    \end{subfigure}
    \begin{subfigure}[t]{0.24\linewidth}
        \centering
        \includegraphics[width=0.95\textwidth]{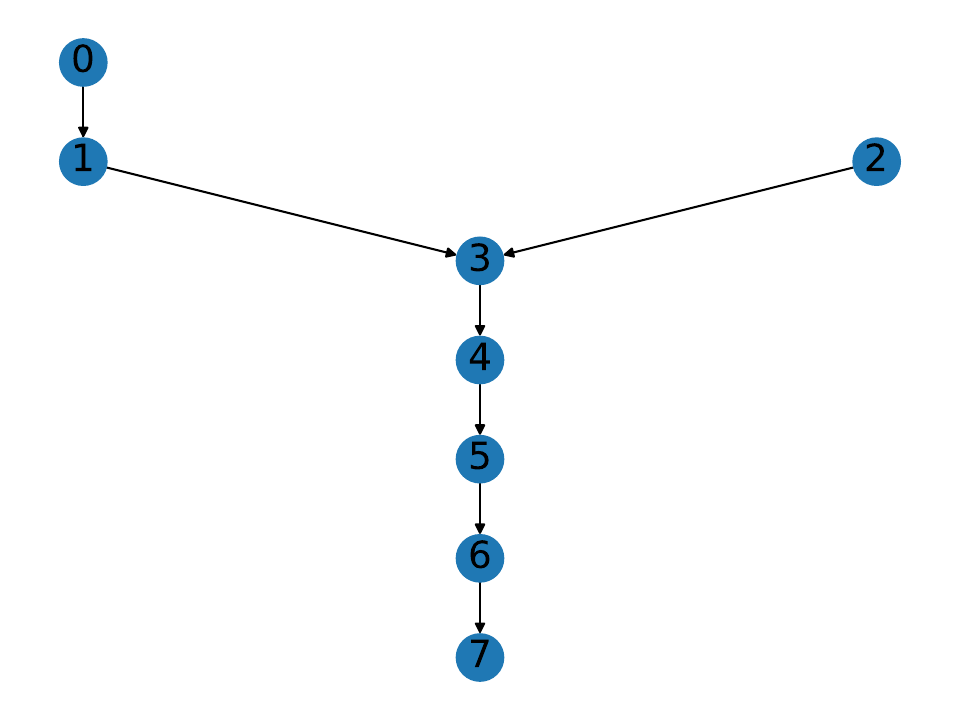}
    \vspace{-1\baselineskip}
        \caption{\centering \small Predicted flow graph from video+text model.}
    \end{subfigure}
    \vspace{-1\baselineskip}
    \caption{An example from YouCookII. The predicted captions are in \textcolor{red}{red} in (a). Video captioning and the text model did not predict the graph correctly, while the video and video+text models did. }
    \vspace{-1\baselineskip}
    \label{fig:youcook}
\end{figure}
\vspace{\baselineskip}
\subsection{Results and Evaluation}
\vspace{-.5\baselineskip}
Table \ref{tab:mmres} shows the results on MM-ReS dataset. For image-only methods, we remove the steps without images during training and evaluation. The flow graphs will change after step removal. When removing steps from the graph, a node is directly deleted if it has no ancestor or descendant. Otherwise, its ancestor is directly connected to its descendant. To enable comparison across modalities, we also evaluate text-only and chain methods on steps with images, denoted by *. We remove nodes without images from text-only model and chain outputs using the above process. Table \ref{tab:youcook} shows the results on YouCookII dataset. 

{\bf Effects of Modalities. } 
For video or image only methods, only Box2Flow-SF surpassed the naive chain baseline in Table \ref{tab:mmres} and \ref{tab:youcook},
showing that directly predicting flow graphs from videos is a challenging problem. We show in our Supplement that the flow graphs predicted by the text model from generated captions do not accurately capture the true structure of the captions. The video captions are inconsistent in ingredients and the scores of manually labeled caption flow graphs would be much lower. Therefore, directly predicting flow graphs from videos is needed. Using text modalities can significantly improve the performance, surpassing the chain baselines. The text model trained on MM-ReS also shows zero-shot ability, achieving high performance on the YouCookII dataset. This is not to be taken for granted, as video clip descriptions are different from formal recipe steps. Furthermore, the step texts in MM-ReS are significantly longer than those in YouCookII. Each step has an average of 32.3 BERT tokens in MM-ReS, while YouCookII only has 14.4 tokens. Yet videos provide complementary information. Methods using both video and text improve upon text-only models. For example, in Figure \ref{fig:recipe_intro2}, \ref{fig:example_flow2}, the text mentions that step 3, "heat up the tortillas", is parallel with steps 2 and 4 as step 3 introduces a new ingredient. However, the video shows the meat is grilled first; then, the tortillas are added while the meat is still grilling and share the same grill, showing the steps are actually sequential. 


{\bf Effects of Bounding Boxes. } Comparing image-only methods and image+text methods in Table \ref{tab:mmres}, video-only methods and video+text methods in Table \ref{tab:youcook}, where step and context features share the same adapter for methods using bounding boxes, the results show using bounding boxes can improve the performance by focusing on the involved ingredients and utensils.

{\bf Effects of Object Detectors, Feature extractors and Interpolation. } Table \ref{tab:mmres} shows SAM masks can further improve the performance from Detectron2 object detectors on COCO. Instruct-pix2pix interpolation improves on the COCO detector more than SAM. 
Table \ref{tab:youcook} shows video-only frame-level SlowFast is better than Densecap features in structures. Including bounding boxes further improves the performance when using SlowFast feature extractor.

In summary, \texttt{Box2flow} can predict the flow graphs effectively and can be used together with different object detectors and feature extractors. Videos can provide information that complements text, and bounding boxes can further improve the effectiveness, even without introducing more parameters in the model. We include more ablation studies, including the effects of context features, SAM mask selection and binary vs soft labels in our Supplement.
\vspace{-2\baselineskip}
\begin{table}[h!]
\small
    \centering
    \caption{MM-ReS results in percentage. The best performance evaluated on all nodes is marked {\bf bold}. * means evaluation on steps with images only.}
    \begin{tabular}{p{1.3cm}p{2cm}p{1.2cm}p{1.3cm}p{1cm}p{1cm}p{1.3cm}p{1.2cm}c}
    \toprule
         Modality&Method&Edge Recall& Edge Precision & Edge F1 &  Recipe  Recall& Recipe Precision & Recipe F1 &MCS\\
         \midrule
         Images &MART\cite{lei2020mart}&80.0&81.0&80.5&81.5&82.5&81.8&77.6 \\
         &VLTinT\cite{yamazaki2023vltint}&  81.8&82.6&82.2&82.2&82.9&82.4& 78.6\\
          &Box2Flow-f&80.5&81.0&80.7&81.5&82.3&81.7&77.7 \\
         &Box2Flow-1 &80.1&80.6&80.3&81.9&82.3&82.0&78.4\\
        \midrule
        Text & Box2Flow &82.5 &83.0&82.8&87.3&87.5&87.4&81.7\\
        &Box2Flow* &82.9 &83.3&83.1&86.1&86.3&86.1&81.7\\
        \midrule
        Images+  & Box2Flow-f & 83.6&83.9&83.7&87.2&87.5&87.3&82.1\\
        Text&  Box2Flow-C&83.7&83.9&83.8&87.3&87.5&87.4&82.8\\
        &Box2Flow-Ci &84.5 &84.7 &84.6 &87.7 &87.9 & 87.8&83.1\\
        & Box2Flow-S &{\bf 84.7}&{\bf 84.9}&{\bf 84.8}&{\bf 87.9}&{\bf 88.1}&{\bf 88.0}&83.3\\
        & Box2Flow-S1 &83.4 &83.6 &83.5 &87.1 &87.3 & 87.2& 82.7\\
        & Box2Flow-Si &84.6 &84.8 &84.7 &87.9 &88.0 & 87.9& {\bf 83.4}\\
        \midrule
        -& chain & 84.1& 84.1& 84.1& 83.4& 83.4&83.4 &78.6\\
        & chain* &85.1 &85.0 & 85.0& 84.5&84.3 &84.3 &80.5\\
        \bottomrule
    \end{tabular}
    \vspace{-2\baselineskip}
    \label{tab:mmres}
\end{table}

\begin{table}[h!]
\small
    \centering
    \vspace{1\baselineskip}
    \caption{YouCookII results in percentage. The best performance is marked {\bf bold}.}
    \begin{tabular}{p{1.3cm}p{2.2cm}p{1.2cm}p{1.3cm}p{1cm}p{1cm}p{1.3cm}p{1.2cm}c}
    \toprule
         Modality&Method&Edge Recall& Edge Precision & Edge F1 &  Recipe  Recall& Recipe Precision & Recipe F1 &MCS\\
         \midrule
         Video &MART\cite{lei2020mart}&74.3&76.0&75.2&77.6&78.9&78.1&70.5\\
         &VLTinT\cite{yamazaki2023vltint}&73.3&75.6&74.4&75.2&76.4&75.7&67.2\\
         &Box2Flow-fD&78.6&79.1&78.8&80.0&80.0&80.0&69.6 \\
        &Box2Flow-fSF&77.0&77.8&77.4&78.9&80.1&79.3&71.8 \\
        &Box2Flow-SF1 & 79.3&79.3&79.3&80.7&80.9&80.7&72.0\\
        &Box2Flow-SF & 80.5&80.3&80.4&81.9&81.5&81.7&72.9\\
        \midrule
        Text & Box2Flow-MMReS & 85.5&85.8&85.6&87.8&88.0&87.9&80.8\\
        \midrule
        Video+  & Box2Flow-fD & 85.8&86.0&85.9&87.9&87.9&87.9&81.7\\
        Text& Box2Flow-fSF &85.5&85.5&85.5&88.1&88.0&88.0&{\bf 81.8}\\
        &Box2Flow-SF& {\bf 86.3}&{\bf 86.5} & {\bf 86.4}& {\bf 88.5}& {\bf 88.7}&{\bf 88.6} &{\bf 81.8}\\
        \midrule
        -& chain& 81.0&80.8&80.9&82.9&82.5&82.7&72.7\\
        \bottomrule
    \end{tabular}
    \vspace{-1\baselineskip}
    \label{tab:youcook}
\end{table}

\vspace{1\baselineskip}
\subsection{Qualitative Results}
\vspace{-.5\baselineskip}
\label{sec:qual}
We show some recipe examples and the flow graph predictions of various methods. The examples are selected based on the largest MCS improvement using video + text from text-only. We include more examples in our Supplement.

Figure \ref{fig:mmres} shows an example from MM-ReS where the text model did not predict the correct graph, but the image+text model did. The shared edges between the ground truth and the text graph are \{(1,2),(3,4)\}; therefore, the precision, recall, and F1 of the recipe are all 2/4=0.5. The maximum common subgraph is 1-2 or 3-4; therefore, MCS=2/5=0.4. For the image+text model, all metrics will be 1. The text model treats step 0 preheat oven and step 1 cook vegetables in a pot as sequential; step 2 mix vegetables and step 3 sprinkle bread crumbs on vegetables as parallel, showing the model did not notice some word details from the long instructions. The images from step 0, 3, 4 are originally missing from the dataset and are interpolated with instruct-pix2pix, marked by \textcolor{blue}{blue} borders. The interpolated images correctly show the oven and baking action in steps 0 and 4 and the bread crumbs in step 3, although missing the vegetables. The image+text model still correctly determined the edge (0,4) and the step 2, 3 should be sequential.

Figure \ref{fig:youcook} shows \texttt{Yummy Pepperoni Pizza Bread} recipe from YouCookII. Video captioning from MART and text model did not predict the correct graph, but video and video+text model did. Both video captioning and the text model predicted chains. The only different edge between the ground truth and the chain is (1,3) in ground truth and (1,2) in the chain; therefore, the precision, recall, and F1 of the recipe are all 6/7=0.875. The maximum common subgraph is the part from 2-7; therefore, MCS=6/8=0.75. For the video and video+text model, all metrics will be 1. MART did not generate the correct ingredients for the first two steps and recognized cheese as butter in step 6. It also generated the impossible action "spread sandwich on bread" in step 3. The captions for steps 4 and 5 are correct. 
The text model predicted the generated captions as a chain even with inconsistent ingredients throughout the recipe.
Meanwhile, the video model and the video+text model correctly predicted step 2 is parallel to the previous steps from the visual clue, correcting the mistake made by the text model.

\vspace{-1\baselineskip}
\section{Conclusion}
\vspace{-0.5\baselineskip}
We have studied the less explored problem, predicting the flow graph from a single procedural video instance. We proposed \texttt{Box2flow} framework, which exploits the bounding boxes and creates a spanning tree from pairwise sequential probabilities. Although a challenging problem, \texttt{Box2flow} can predict the flow graphs effectively. 
This also opens up possible future research directions: predicting the flow graphs more effectively from video or image-only features and exploring their utility in downstream applications, like more structured video captioning and planning.


%
%
\bibliographystyle{splncs04}
\bibliography{reference}
\end{document}